\documentclass{article} % For LaTeX2e
\usepackage{sdnet,times}

\usepackage{amsmath,amsfonts,bm}

\def\eqref#1{equation~\ref{#1}}
\def\1{\bm{1}}

\DeclareMathAlphabet{\mathsfit}{\encodingdefault}{\sfdefault}{m}{sl}
\SetMathAlphabet{\mathsfit}{bold}{\encodingdefault}{\sfdefault}{bx}{n}

\usepackage{hyperref}
\usepackage{url}
\usepackage{amssymb}
\usepackage{wrapfig,lipsum,booktabs}
\usepackage{amsmath}
\usepackage{cleveref}
\usepackage{booktabs}
\usepackage{bm}
\usepackage{algorithm}
\usepackage{algpseudocode}
\usepackage{graphicx}
\usepackage{nccmath}
\usepackage{caption}
\usepackage{makecell}
\usepackage{subcaption}
\usepackage{tabularx}
\newcommand{\B}[1] {\boldsymbol{#1}}
\def\bA{{\B{A}}}
\def\bB{{\B{B}}}
\def\bC{{\B{C}}}

\def\bh{{\B{h}}}
\def\bm{{\B{m}}}

\def\bs{{\B{s}}}
\def\bt{{\B{t}}}
\def\bu{{\B{u}}}
\def\bv{{\B{v}}}
\def\bw{{\B{w}}}

\def\by{{\B{y}}}

\DeclareMathOperator*{\relu}{ReLU}

\DeclareMathOperator{\gru}{GRU}

\DeclareMathOperator{\BiLSTM}{\mbox{BiLSTM}}
\DeclareMathOperator{\bert}{\mbox{BERT}}
\DeclareMathOperator{\glove}{GloVe}

\title{SDNet: Contextualized Attention-based Deep Network for Conversational Question Answering}

\author{Chenguang Zhu$^{1}$, Michael Zeng$^{1}$, Xuedong Huang$^{1}$\\
$^1$ Microsoft Speech and Dialogue Research Group, Redmond, WA 98052, USA\\
\texttt{\{chezhu, nzeng, xdh\}@microsoft.com} \\
}

\iclrfinalcopy
\begin{document}
\maketitle

\begin{abstract}
Conversational question answering (CQA) is a novel QA task that requires understanding of dialogue context. Different from traditional single-turn machine reading comprehension (MRC) tasks, CQA includes passage comprehension, coreference resolution, and contextual understanding. In this paper, we propose an innovated contextualized attention-based deep neural network, SDNet, to fuse context into traditional MRC models. Our model 
leverages both inter-attention and self-attention to comprehend conversation context and extract relevant information from passage. Furthermore, we demonstrated a novel method to integrate the latest BERT contextual model. Empirical results show the effectiveness of our model, which sets the new state of the art result in CoQA leaderboard, outperforming the previous best model by 1.6\% $F_1$. Our ensemble model further improves the result by 2.7\% $F_1$.
\end{abstract}

\section{Introduction}
Traditional machine reading comprehension (MRC) tasks share the single-turn setting of answering a single question related to a passage. There is usually no connection between different questions and answers to the same passage. However, the most natural way humans seek answers is via conversation, which carries over context through the dialogue flow. 

To incorporate conversation into reading comprehension, recently there are several public datasets that evaluate QA model's efficacy in conversational setting, such as CoQA \citep{coqa}, QuAC \citep{quac} and QBLink \citep{qblink}. In these datasets, to generate correct responses, models are required to fully understand the given passage as well as the context of previous questions and answers. Thus, traditional neural MRC models are not suitable to be directly applied to this scenario. Existing approaches to conversational QA tasks include BiDAF++ \citep{bidafplusplus}, FlowQA \citep{flowqa}, DrQA+PGNet \citep{coqa}, which all try to find the optimal answer span given the passage and dialogue history.

In this paper, we propose SDNet, a contextual attention-based deep neural network for the task of conversational question answering. Our network stems from machine reading comprehension models, but has several unique characteristics to tackle contextual understanding during conversation. Firstly, we apply both inter-attention and self-attention on passage and question to obtain a more effective understanding of the passage and dialogue history.
Secondly, SDNet leverages the latest breakthrough in NLP: BERT contextual embedding \citep{bert}. Different from the canonical way of appending a thin layer after BERT structure according to \citep{bert}, we innovatively employed a weighted sum of BERT layer outputs, with locked BERT parameters. Thirdly, we prepend previous rounds of questions and answers to the current question to incorporate contextual information. Empirical results show that each of these components has substantial gains in prediction accuracy.

We evaluated SDNet on CoQA dataset, which improves the previous state-of-the-art model's result by 1.6\% (from 75.0\% to 76.6\%) overall $F_1$ score. The ensemble model further increase the $F_1$ score to $79.3\%$. Moreover, SDNet is the first model ever to pass $80\%$ on CoQA's in-domain dataset. 

\section{Approach}
In this section, we propose the neural model, SDNet, for the conversational question answering task, which is formulated as follows. Given a passage $\mathcal{C}$, and history question and answer utterances $Q_1, A_1, Q_2, A_2, ..., Q_{k-1}, A_{k-1}$, the task is to generate response $A_k$ given the latest question $Q_k$. The response is dependent on both the passage and history utterances.

To incorporate conversation history into response generation, we employ the idea from DrQA+PGNet \citep{coqa} to prepend the latest $N$ rounds of utterances to the current question $Q_k$ . The problem is then converted into a machine reading comprehension task. In other words, the reformulate question is $\mathcal{Q}_k=\{Q_{k-N}; A_{k-N}; ..., Q_{k-1}; A_{k-1}; Q_k\}$. To differentiate between question and answering, we add symbol $\langle Q \rangle$ before each question and $\langle A \rangle$ before each answer in the experiment.

\subsection{Model Overview}

\begin{figure*}[t]
\hspace*{-1.25cm}
\includegraphics[scale=0.65,trim=0cm 0cm 0cm 4cm]{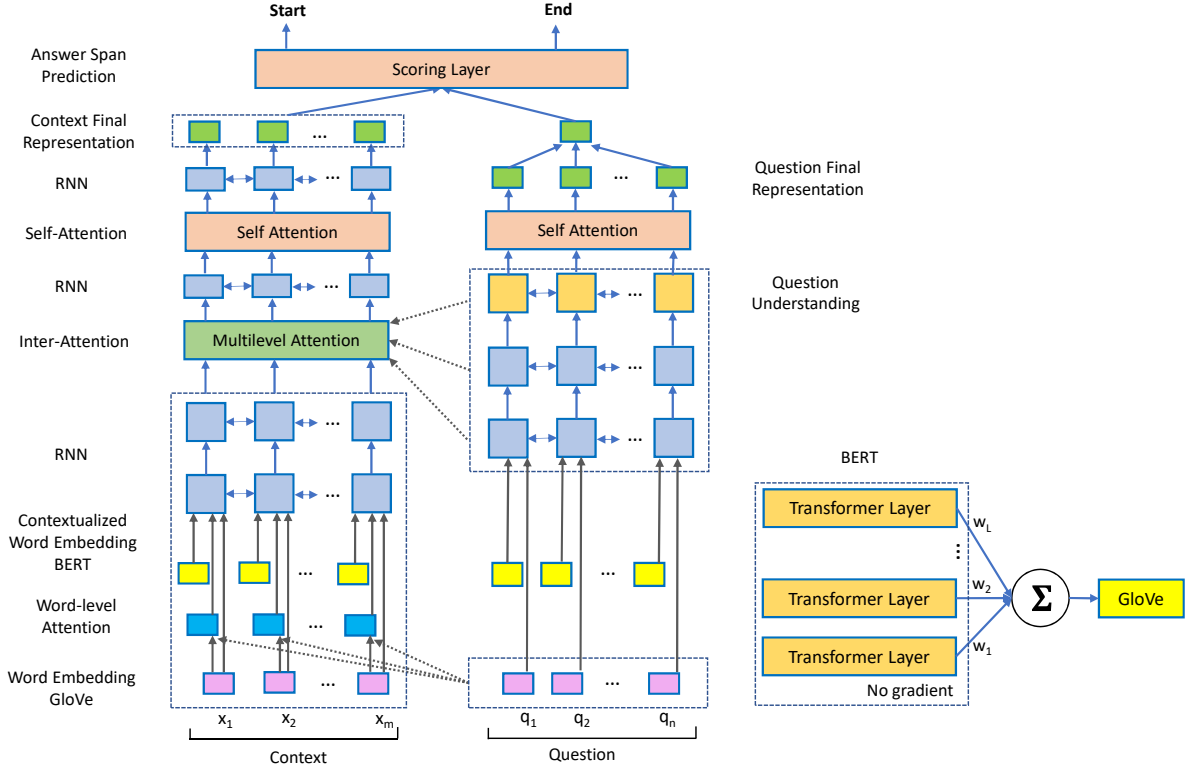}
\vspace{-5\baselineskip}
\caption{SDNet model structure.}
\label{fig:model}
\end{figure*}

\textit{Encoding layer} encodes each token in passage and question into a fixed-length vector, which includes both word embeddings and contextualized embeddings. For contextualized embedding, we utilize the latest result from BERT \citep{bert}. Different from previous work, we fix the parameters in BERT model and use the linear combination of embeddings from different layers in BERT.

\textit{Integration layer} uses multi-layer recurrent neural networks (RNN) to capture contextual information within passage and question. To characterize the relationship between passage and question, we conduct word-level attention from question to passage both before and after the RNNs. We employ the idea of history-of-word from FusionNet \citep{fusionnet} to reduce the dimension of output hidden vectors. Furthermore, we conduct self-attention to extract relationship between words at different positions of context and question.

\textit{Output layer} computes the final answer span. It uses attention to condense the question into a fixed-length vector, which is then used in a bilinear projection to obtain the probability that the answer should start and end at each position. 

An illustration of our model SDNet is in \Cref{fig:model}.

\subsection{Encoding layer} 
We use 300-dim GloVe \citep{pennington2014glove} embedding and contextualized embedding for each word in context and question. We employ BERT \citep{bert} as contextualized embedding. Instead of adding a scoring layer to BERT structure as proposed in \citep{bert}, we use the transformer output from BERT as contextualized embedding in our encoding layer. BERT generates $L$ layers of hidden states for all BPE tokens \citep{bpe} in a sentence/passage and we employ a weighted sum of these hidden states to obtain contextualized embedding. Furthermore, we lock BERT's internal weights, setting their gradients to zero. In ablation studies, we will show that this weighted sum and weight-locking mechanism can significantly boost the model's performance.

In detail, suppose a word $w$ is tokenized to $s$ BPE tokens $w=\{b_1, b_2, ..., b_s\}$, and BERT generates $L$ hidden states for each BPE token, $\mathbf{h^l_t}, 1\leq l \leq L, 1\leq t \leq s$. The contextual embedding $\bert_w$ for word $w$ is then a per-layer weighted sum of average BERT embedding, with weights $\alpha_1, ..., \alpha_L$.

$$\bert_w = \sum_{l=1}^L \alpha_l \frac{\sum_{t=1}^s \mathbf{h}^l_t}{s}
$$

\subsection{Integration layer}
\textbf{Word-level Inter-Attention.} We conduct attention from question to context (passage) based on GloVe word embeddings. Suppose the context word embeddings are $\{\bh^C_1, ..., \bh^C_m\}\subset \mathbb{R}^d$, and the question word embeddings are $\{\bh^Q_1, ..., \bh^Q_n\}\subset \mathbb{R}^d$. Then the attended vectors from question to context are $\{\hat{\bh}^C_1, ..., \hat{\bh}^C_m\}$, defined as,
$$S_{ij} = \relu(Uh^C_i)D\relu(Uh^Q_j),$$ 
$$\alpha_{ij} \propto  {exp(S_{ij})},$$
$$\hat{\bh}^C_i = \sum_j \alpha_{ij}\bh^Q_j,$$

where $D\in \mathbb{R}^{k\times k}$ is a diagonal matrix and $U\in \mathbb{R}^{d\times k}$, $k$ is the attention hidden size. 

To simplify notation, we define the attention function above as $\mbox{Attn}(\bA, \bB, \bC)$, meaning we compute the attention score $\alpha_{ij}$ based on two sets of vectors $\bA$ and $\bB$, and use that to linearly combine vector set $\bC$. So the word-level attention above can be simplified as $\mbox{Attn}(\{\bh^C_i\}_{i=1}^m, \{\bh^Q_i\}_{i=1}^n\}, \{\bh^Q_i\}_{i=1}^n\})$.

For each context word in $\mathcal{C}$, we also include a feature vector $f_w$ including 12-dim POS embedding, 8-dim NER embedding, a 3-dim exact matching vector $em_i$ indicating whether each context word appears in the question, and a normalized term frequency, following the approach in DrQA \citep{drqa}.

Therefore, the input vector for each context word is $\Tilde{\bw}_i^C=[\glove(w_i^C); \bert_{w_i^C}; \hat{\bh}^C_i; f_{w_i^C}]$; the input vector for each question word is $\Tilde{\bw}_i^Q=[\glove(w_i^Q); \bert_{w_i^Q}]$.

\textbf{RNN.} In this component, we use two separate bidirectional RNNs (BiLSTMs \citep{lstm}) to form the contextualized understanding for $\mathcal{C}$ and $\mathcal{Q}$.
$$
\bh_1^{C,k}, ..., \bh_m^{C,k} = \BiLSTM{(\bh_1^{C,k-1}, ..., \bh_m^{C,k-1})},
$$
$$
\bh_1^{Q,k}, ..., \bh_n^{Q,k} = \BiLSTM{(\bh_1^{Q,k-1}, ..., \bh_n^{Q,k-1})},
$$
$$
\bh_i^{C,0} = \Tilde{\bw}_i^C, \bh_i^{Q,0} = \Tilde{\bw}_i^Q,
$$
where $1\leq k \leq K$ and $K$ is the number of RNN layers. We use variational dropout \citep{vd} for input vector to each layer of RNN, i.e. the dropout mask is shared over different timesteps.

\textbf{Question Understanding.} For each question word in $\mathcal{Q}$, we employ one more layer of RNN to generate a higher level of understanding of the question.
$$
\bh_1^{Q,K+1}, ..., \bh_n^{Q,K+1} = \BiLSTM{(\bh_1^{Q}, ..., \bh_n^{Q})},
$$
$$
\bh_i^{Q} = [\bh_i^{Q,1};...;\bh_i^{Q,K}]
$$

\textbf{Self-Attention on Question.} As the question has integrated previous utterances, the model needs to directly relate previously mentioned concept with the current question. This is helpful for concept carry-over and coreference resolution. We thus employ self-attention on question. The formula is the same as word-level attention, except that we are attending a question to itself: $\{\bu_i^Q\}_{i=1}^n=\mbox{Attn}(\{\bh_i^{Q,K+1}\}_{i=1}^n, \{\bh_i^{Q,K+1}\}_{i=1}^n, \{\bh_i^{Q,K+1}\}_{i=1}^n)$. The \textbf{final question representation} is thus $\{\bu_i^Q\}_{i=1}^n$.

\textbf{Multilevel Inter-Attention.} After multiple layers of RNN extract different levels of understanding of each word, we conduct multilevel attention from question to context based on all layers of generated representations.

However, the aggregated dimensions can be very large, which is computationally inefficient. We thus leverage the history-of-word idea from FusionNet \citep{fusionnet}: we use all previous levels to compute attentions scores, but only linearly combine RNN outputs.

In detail, we conduct $K+1$ times of multilevel attention from each RNN layer output of question to context.

\begin{align*}
\{\bm_i^{(k),C}\}_{i=1}^m=\mbox{Attn}(\{\mbox{HoW}_i^C\}_{i=1}^m, \{\mbox{HoW}_i^Q\}_{i=1}^n,\{\bh_i^{Q,k}\}_{i=1}^n), 1\leq k \leq K+1
\end{align*}

where history-of-word vectors are defined as
$$\mbox{HoW}_i^C = [\glove(w_i^C); \bert_{w_i^C}; \bh_i^{C,1}; ..., \bh_i^{C,k}],$$
$$\mbox{HoW}_i^Q = [\glove(w_i^Q); \bert_{w_i^Q}; \bh_i^{Q,1}; ..., \bh_i^{Q,k}].$$

An additional RNN layer is applied to obtain the contextualized representation $\bv_i^C$ for each word in $\mathcal{C}$.
$$
\by_i^C = [\bh_i^{C,1}; ...; \bh_i^{C,k}; \bm_i^{(1),C}; ...; \bm_i^{(K+1),C}],
$$
$$
\bv_1^{C}, ..., \bv_m^{C} = \BiLSTM{(\by_1^{C}, ..., \by_n^{C})},
$$
%\textbf{Flow component.} We employ the concept of flow \citep{flowqa}. As we group questions and answers from the same passage in order in one batch, flow component is to apply a forward RNN to fuse information from the same context word in previous rounds to later round. This can facilitate the understanding of the context as dialogue flows forward.

%Specifically, if a passage has $A$ rounds of questions, the batch will contain $A$ copies of context, each corresponding to a current question. Suppose $\bp_{i,a}^C$ is the embedding of the $i$-th context word in the $a$-th context, $1\leq a \leq A$:
%\begin{align*}
%\bp_{i,a}^C=&[\glove(w_{i,a}^C); \bert_{w_{i,a}^C}; \bh_{i,a}^{C,1}; ...; \bh_{i,a}^{C,k}; \\
% &\bm_{i,a}^{(1),Q}; ...; \bm_{i,a}^{(K+1),Q}; \bv_{i,a}^C]
%\end{align*}

%Then, the forward flow RNN runs over the embedding for each word in all $M$ rounds:
%$$\{\flow_{i,a}^C\}_{a=1}^A = \ForwardLSTM{(\bp_{i,1}^C, \bp_{i,2}^C, ..., \bp_{i,A}^C)}$$
 
%Without ambiguation bewteen different instances in the same batch, we denote the flow output by $\flow_i^C$.

\textbf{Self Attention on Context.} Similar to questions, we conduct self attention on context to establish direct correlations between all pairs of words in $\mathcal{C}$. Again, we use the history of word concept to reduce the output dimension by linearly combining $\bv_i^C$. 

\begin{align*}
\bs_i^C  = &[\glove(w_i^C); \bert_{w_i^C}; \bh_i^{C,1}; ...; \bh_i^{C,k}; \bm_i^{(1),Q}; ...; \bm_i^{(K+1),Q}; \bv_i^C]
\end{align*}

$$\{\Tilde{\bv}_i^C\}_{i=1}^m=\mbox{Attn}(\{\bs_i^C\}_{i=1}^m, \{\bs_i^C\}_{i=1}^m, \{\bv_i^C\}_{i=1}^m)$$

The self-attention is followed by an additional layer of RNN to generate the \textbf{final representation of context}:
$$\{\bu_i^C\}_{i=1}^m = \BiLSTM{([\bv_1^C; \Tilde{\bv}_1^C], ..., [\bv_m^C; \Tilde{\bv}_m^C])}$$

%\textbf{Final Representation of Context.} A final flow component is used to generate the final representation of context:

%$$\{\bu_{i,a}^C\}_{a=1}^A = \ForwardLSTM{([\br_{i,1}^C, \br_{i,2}^C, ..., \br_{i,A}^C])}$$

%Without ambiguation bewteen different instances in the same batch, we denote the final representation of context by $\{\bu_i^C\}_{i=1}^m$.

\subsection{Output layer}
\textbf{Generating Answer Span.} This component is to generate two scores for each context word corresponding to the probability that the answer starts and ends at this word, respectively.

Firstly, we condense the question representation into one vector: $\bu^Q=\sum_i{\beta_i}\bu_i^Q$, where $\beta_i\propto{\exp{(\bw^T\bu_i^Q)}}$ and $\bw$ is a parametrized vector.

Secondly, we compute the probability that the answer span should start at the $i$-th word:
$$P_i^S\propto{\exp{((\bu^Q)^TW_S\bu_i^C)}},$$
where $W_S$ is a parametrized matrix. We further fuse the start-position probability into the computation of end-position probability via a GRU, $\bt^Q = \gru{(\bu^Q, \sum_i P_i^S\bu_i^C)}$. Thus, the probability that the answer span should end at the $i$-th word is:
$$P_i^E\propto{\exp{((\bt^Q)^TW_E\bu_i^C)}},$$
where $W_E$ is another parametrized matrix.

For CoQA dataset, the answer could be affirmation ``yes'', negation ``no'' or no answer ``unknown''. We separately generate three probabilities corresponding to these three scenarios, $P_Y, P_N, P_U$, respectively. For instance, to generate the probability that the answer is ``yes'', $P_Y$, we use:
$$P_i^{Y}\propto{\exp{((\bu^Q)^T W_{Y}\bu_i^C})},$$
$$P_{Y} = (\sum_i P_i^{Y}\bu_i^C)^T\bw_{Y},$$
where $W_Y$ and $\bw_Y$ are parametrized matrix and vector, respectively.

\textbf{Training.} For training, we use all questions/answers for one passage as a batch. The goal is to maximize the probability of the ground-truth answer, including span start/end position, affirmation, negation and no-answer situations. Equivalently, we minimize the negative log-likelihood function $\mathcal{L}$:
\begin{align*}
\mathcal{L} =& \sum_k I^S_k(\mbox{log}(P^S_{i_k^s}) + \mbox{log}(P^E_{i_k^e})) + I^Y_k\mbox{log}P^Y_k+I^N_k\mbox{log}P^N_k + I^U_k\mbox{log}P^U_k,
\end{align*}
where $i_k^s$ and $i_k^e$ are the ground-truth span start and end position for the $k$-th question. $I^S_k, I^Y_k, I^N_k, I^U_k$ indicate whether the $k$-th ground-truth answer is a passage span, ``yes'', ``no'' and ``unknown'', respectively. More implementation details are in Appendix.

\textbf{Prediction.} During inference, we pick the largest span/yes/no/unknown probability. The span is constrained to have a maximum length of 15.

\section{Experiments}
\label{exp}
We evaluated our model on CoQA \citep{coqa}, a large-scale conversational question answering dataset. In CoQA, many questions require understanding of both the passage and previous questions and answers, which poses challenge to conventional machine reading models. \Cref{table:coqa} summarizes the domain distribution in CoQA. As shown, CoQA contains passages from multiple domains, and the average number of question answering turns is more than 15 per passage. Many questions require contextual understanding to generate the correct answer.

\begin{table}[t]
\centering
\caption{Domain distribution in CoQA dataset.}\label{table:coqa}
\vspace{-0.5\baselineskip}
\begin{tabular}{lcc}
\toprule
Domain & \#Passage & \#QA turn\\ \midrule
Child Story & 750 & 14.0\\
Literature & 1,815 & 15.6\\
Mid/High Sc. & 1,911 & 15.0\\
News & 1,902 & 15.1\\
Wikipedia & 1,821 &  15.4 \\ \midrule 
& Out of domain & \\ \midrule 
Science & 100 &  15.3\\
Reddit & 100 & 16.6\\ \midrule
Total & 8,399 & 15.2\\
\bottomrule

\end{tabular}
\end{table}

For each in-domain dataset, 100 passages are in the development set, and 100 passages are in the test set. The rest in-domain dataset are in the training set. The test set also includes all of the out-of-domain passages.

\textbf{Baseline models and metrics.}
We compare SDNet with the following baseline models: PGNet (Seq2Seq with copy mechanism) \citep{pgnet}, DrQA \citep{drqa}, DrQA+PGNet \citep{coqa}, BiDAF++ \citep{bidafplusplus} and FlowQA \citep{flowqa}. Aligned with the official leaderboard, we use $F_1$ as the evaluation metric, which is the harmonic mean of precision and recall at word level between the predicted answer and ground truth.\footnote{According to official evaluation of CoQA, when there are more than one ground-truth answers, the final score is the average of max $F_1$ against all-but-one ground-truth answers.}

\begin{table*}[t]
\centering
\caption{Model and human performance (\% in F1 score) on the CoQA test set.} 
\label{table:mainresult}
\vspace{-0.5\baselineskip}
\setlength{\tabcolsep}{2pt}
\begin{tabular}{l|ccccccc|c}
\toprule
 & Child. & Liter. & Mid-High. & News & Wiki & Reddit & Science & Overall \\
 \midrule
 \midrule
PGNet & 49.0 & 43.3 & 47.5 & 47.5 & 45.1 & 38.6 & 38.1 & 44.1 \\
DrQA & 46.7 & 53.9 & 54.1 & 57.8 & 59.4 & 45.0 & 51.0 & 52.6 \\
DrQA+PGNet & 64.2 & 63.7 & 67.1 & 68.3 & 71.4 & 57.8 & 63.1 & 65.1 \\
BiDAF++ & 66.5 & 65.7 & 70.2 & 71.6 & 72.6 & 60.8 & 67.1 & 67.8 \\
FlowQA & 73.7 & 71.6 & 76.8 & 79.0 & 80.2 & 67.8 & 76.1 & 75.0 \\
SDNet (single) & \textbf{75.4} & \textbf{73.9} & \textbf{77.1} & \textbf{80.3} & \textbf{83.1} & \textbf{69.8} & \textbf{76.8} & \textbf{76.6} \\
SDNet (ensemble) & \textbf{78.7} & \textbf{77.1} & \textbf{80.2} & \textbf{81.9} & \textbf{85.2} & \textbf{72.3} & \textbf{79.7} & \textbf{79.3} \\
\midrule
Human & 90.2 & 88.4 & 89.8 & 88.6 & 89.9 & 86.7 & 88.1 & 88.8 
\end{tabular}
\end{table*}

\textbf{Results.}  \Cref{table:mainresult} report the performance of SDNet and baseline models.\footnote{Result was taken from official CoQA leaderboard on Nov. 30, 2018.} As shown, SDNet achieves significantly better results than baseline models. In detail, the single SDNet model improves overall $F_1$ by 1.6\%, compared with previous state-of-art model on CoQA, FlowQA. Ensemble SDNet model further improves overall $F_1$ score by 2.7\%, and it's the first model to achieve over 80\% $F_1$ score on in-domain datasets (80.7\%).

\begin{figure*}[t]
\centering
\includegraphics[width=\textwidth]{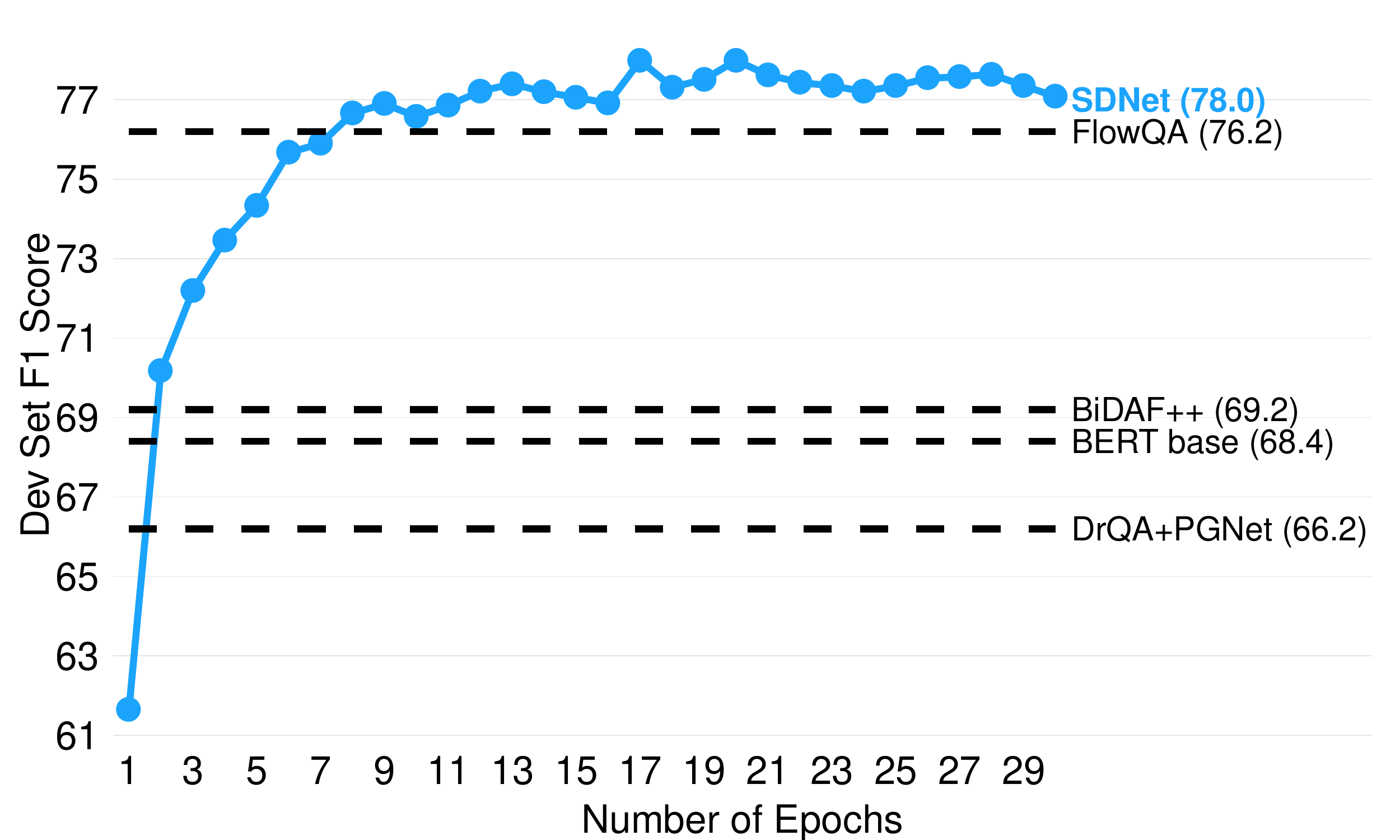}
\vspace{-1\baselineskip}
\caption{$F_1$ score on CoQA dev set over training epochs. For BERT base model, as there is no associated paper, we use the number on test set from the leaderboard.}
\label{fig:epoch}
\end{figure*}

\Cref{fig:epoch} shows the $F_1$ score on development set over epochs. As seen, SDNet overpasses all but one baseline models after the second epoch, and achieves state-of-the-art results only after 8 epochs.  

\textbf{Ablation Studies.} We conduct ablation studies on SDNet model and display the results in \Cref{table:ablation}. The results show that removing BERT can reduce the $F_1$ score on development set by $7.15\%$. Our proposed weight sum of per-layer output from BERT is crucial, which can boost the performance by $1.75\%$, compared with using only last layer's output. This shows that the output from each layer in BERT is useful in downstream tasks. This technique can also be applied to other NLP tasks. Using BERT-base instead of BERT-large pretrained model hurts the $F_1$ score by $2.61\%$, which manifests the superiority of BERT-large model. Variational dropout and self attention can each improve the performance by 0.24\% and 0.75\%, respectively.

\begin{table}[t]
\centering
\caption{Ablation study of SDNet on CoQA development dataset.}
\label{table:ablation}
\begin{tabular}{lc}
\toprule
Model & $F_1$ \\ 
\midrule
\midrule
SDNet & 77.99 \\
\quad --Variational dropout & 77.75 \\
\quad --Question self attention & 77.24 \\
\quad Using last layer of BERT output &  \\
\quad (no weighted sum) & 76.24 \\
\quad BERT-base & 75.38 \\
\quad --BERT & 70.84 \\
%\quad --BERT lock weights & 64.15 \\
\bottomrule
\end{tabular}
\end{table}

\textbf{Contextual history.} In SDNet, we utilize conversation history via prepending the current question with previous $N$ rounds of questions and ground-truth answers. We experimented the effect of $N$ and show the result in \Cref{table:context}. Excluding dialogue history ($N=0$) can reduce the $F_1$ score by as much as $8.56\%$, showing the importance of contextual information in conversational QA task.
The performance of our model peaks when $N=2$, which was used in the final SDNet model.

\begin{table}[h]
\centering
\caption{Performance of SDNet on development set when prepending different number of history questions and answers to the question. The model uses BERT-Large contextual embedding and fixes BERT's weights.}
\label{table:context}
\begin{tabular}{ccc}
\toprule
\#previous QA rounds $N$ & $F_1$ \\ \midrule
0 & 69.43 \\
1 & 76.70\\
2 & \textbf{77.99}\\
3 & 77.39\\
\bottomrule
\end{tabular}
\end{table}

\section{Conclusions}
\label{conclusion}
In this paper, we propose a novel contextual attention-based deep neural network, SDNet, to tackle conversational question answering task. By leveraging inter-attention and self-attention on passage and conversation history, the model is able to comprehend dialogue flow and fuse it with the digestion of passage content. Furthermore, we incorporate the latest breakthrough in NLP, BERT, and leverage it in an innovative way. SDNet achieves superior results over previous approaches. On the public dataset CoQA, SDNet outperforms previous state-of-the-art model by 1.6\% in overall $F_1$ metric.

Our future work is to apply this model to open-domain multiturn QA problem with large corpus or knowledge base, where the target passage may not be directly available. This will be an even more realistic setting to human question answering.

\bibliography{sdnet}
\bibliographystyle{sdnet}

\appendix \label{sec:appendix}
\section{Implementation Details}
We use spaCy for tokenization. As BERT use BPE as the tokenizer, we did BPE tokenization for each token generated by spaCy. In case a token in spaCy corresponds to multiple BPE sub-tokens, we average the BERT embeddings of these BPE sub-tokens as the embedding for the token. We fix the BERT weights and use the BERT-Large-Uncased model.

During training, we use a dropout rate of 0.4 for BERT layer outputs and 0.3 for other layers. We use variational dropout \citep{vd}, which shares the dropout mask over timesteps in RNN. We batch the data according to passages, so all questions and answers from the same passage make one batch. 

We use Adamax \citep{adamax} as the optimizer, with a learning rate of $\alpha=0.002, \beta=(0.9, 0.999)$ and $\epsilon=10^{-8}$. We train the model using 30 epochs, with each epoch going over the data once. We clip the gradient at length $10$.

The word-level attention has a hidden size of 300. The flow module has a hidden size of 300. The question self attention has a hidden size of 300. The RNN for both question and context has $K=2$ layers and each layer has a hidden size of 125. The multilevel attention from question to context has a hidden size of 250. The context self attention has a hidden size of 250. The final layer of RNN for context has a hidden size of 125.
\end{document}